\documentclass[pdflatex,sn-mathphys-num]{sn-jnl}
\usepackage{graphicx}
\usepackage{multirow}
\usepackage{amsmath,amssymb,amsfonts}
\usepackage{amsthm}
\usepackage{mathrsfs}
\usepackage[title]{appendix}
\usepackage{textcomp}
\usepackage{manyfoot}
\usepackage{booktabs}
\usepackage{algorithm}
\usepackage{algorithmicx}
\usepackage{algpseudocode}
\usepackage{listings}

\usepackage{array}
\usepackage{xltabular} 
\usepackage[table]{xcolor}

\usepackage{arabtex}
\usepackage{utf8}

\usepackage{CJKutf8}
\newcommand{\textchinese}[1]{\begin{CJK}{UTF8}{gbsn}#1\end{CJK}}

\begin{document}

\setcode{utf8}
\title[Article Title]{Grounding Arabic LLMs in the Doha Historical Dictionary: Retrieval-Augmented Understanding of Qur’an and Hadith}

\author[1]{\fnm{Somaya} \sur{Eltanbouly}}\email{seltanbouly@hbku.edu.qa}

\author[1]{\fnm{Samer} \sur{Rashwani}}\email{srashwani@hbku.edu.qa}

\affil[1]{\orgdiv{College of Islamic Studies}, \orgname{Hamad bin Khalifa University}, \orgaddress{\city{Doha}, \country{Qatar}}}

\abstract{Large language models (LLMs) have achieved remarkable progress in many language tasks, yet they continue to struggle with complex historical and religious Arabic texts such as the Qur’an and Hadith. To address this limitation, we develop a retrieval-augmented generation (RAG) framework grounded in diachronic lexicographic knowledge. Unlike prior RAG systems that rely on general-purpose corpora, our approach retrieves evidence from the Doha Historical Dictionary of Arabic (DHDA), a large-scale resource documenting the historical development of Arabic vocabulary. The proposed pipeline combines hybrid retrieval with an intent-based routing mechanism to provide LLMs with precise, contextually relevant historical information. Our experiments show that this approach improves the accuracy of Arabic-native LLMs, including Fanar and ALLaM, to over 85\%, substantially reducing the performance gap with Gemini, a proprietary large-scale model. Gemini also serves as an LLM-as-a-judge system for automatic evaluation in our experiments. The automated judgments were verified through human evaluation, demonstrating high agreement ($\kappa = 0.87$). An error analysis further highlights key linguistic challenges, including diacritics and compound expressions. These findings demonstrate the value of integrating diachronic lexicographic resources into retrieval-augmented generation frameworks to enhance Arabic language understanding, particularly for historical and religious texts. The code and resources are publicly available at:
\url{https://github.com/somayaeltanbouly/Doha-Dictionary-RAG}.}

\keywords{Retrieval-Augmented Generation, Large Language Models, Historical Arabic, Qur’an and Hadith}

\maketitle

\section{Introduction}\label{sec1}

Large Language Models (LLMs) and Retrieval Augmented Generation (RAG) have significantly advanced various tasks in Natural Language Processing (NLP), including question answering, summarization, and knowledge support. However, most progress has focused on high-resource languages like English, while Arabic remains a low-resource language, particularly in historical and religious contexts \cite{llm_arabic_survey}. 

Classical and religious texts pose unique challenges, as they contain archaic vocabulary absent from modern sources, terms whose meanings have shifted over time, and grammatical structures and styles that differ from present-day norms. Moreover, they carry rich cultural and interpretive layers that must be taken into account \cite{Ara_LLMS, inclusive_arabic_LLM}. These characteristics render standard LLMs, often trained on contemporary web text, prone to anachronistic errors and factual inaccuracies when interpreting such rich materials. Thus, without explicitly addressing these challenges, AI systems risk misinterpreting or oversimplifying such complex texts.

To address these challenges, several Arabic-specific pretrained models have been developed with a focus on better representing the linguistic diversity of Arabic, including Classical Arabic. Early encoder-based approaches such as CAMeL-BERT \cite{camelbert} and CL-AraBERT \cite{clarabert} demonstrated improved performance on tasks that rely on classical registers, including Qur’anic question answering and retrieval. With the advent of large generative models, the Arabic NLP landscape has expanded significantly. Notable contributions include JAIS and Falcon, both released in 2023 \cite{jais_paper, falcon_paper}, followed by more recent efforts such as SILMA \cite{silma-9b-2024}, ALLaM \cite{bari2025allam}, and Fanar \cite{fanarllm2025}. More recently, mid-sized Arabic LLMs have also been released, such as JAIS V2 (70B) \cite{jais2_2025} and Karnak (40B) \cite{karnak-40b}. These models report competitive results on common Arabic benchmarks.

Despite these advancements, current Arabic LLMs still face limitations when applied to linguistically demanding tasks. Their comparatively smaller scale and the scarcity of high-quality, domain-specific corpora mean that performance often falls short on complex varieties such as Classical Arabic or highly interpretive domains like poetry and the interpretation of religious texts \cite{fann_flop, Ara_LLMS}. This limitation is further exacerbated by the inherent variation across Arabic dialects, Modern Standard Arabic, and Classical Arabic, which complicates generalization. Thus, while Arabic-focused LLMs represent a substantial step forward, further progress is required to enable them to work effectively with complex Arabic texts such as those of historical and religious significance.

To improve LLMs’ ability to understand complex Arabic texts, models need to be trained with or augmented by external knowledge sources that capture Arabic lexical information. For example, in \cite{al-rasheed-etal-2025-evaluating}, the authors used a modern Arabic dictionary as an external knowledge source, enabling the LLM to answer questions related to Arabic lexical meanings. However, their work focused exclusively on Modern Standard Arabic; consequently, the models’ capability to address questions involving historical Arabic remains largely unexplored. Accordingly, the present study aims to enhance LLMs’ capacity to understand and interpret historical Arabic, including word meanings and other details like their development and origins. Importantly, addressing historical Arabic requires a resource that not only defines words but also traces their semantic evolution through time with evidentiary citations. Modern dictionaries lack this crucial diachronic dimension.

To address these limitations, we propose a RAG approach that utilizes the rich, structured resources of the Doha Historical Dictionary of Arabic\footnote{\url{https://www.dohadictionary.org/}}. By incorporating the dictionary’s detailed lexical information as external knowledge, our framework enables LLMs to more accurately interpret complex Arabic texts, particularly those of historical and religious significance. Accordingly, the objectives of this research are:
\begin{enumerate}
    \item To design and evaluate a retrieval pipeline optimized for the rich, structured data of a historical dictionary, capable of matching complex semantic and historical queries.
    \item To benchmark the performance of Arabic-native LLMs in a RAG framework grounded in a historical Arabic dictionary, specifically testing their ability to synthesize accurate answers from retrieved historical lexical data.
    \item To demonstrate how augmenting LLMs with historical lexicographical knowledge significantly enhances their capacity for nuanced interpretation of foundational texts like the Quran and Hadith
\end{enumerate}

The remainder of this paper is organized as follows. Section \ref{lr} reviews related research. Section \ref{methodology} details the proposed RAG methodology, covering preprocessing, retrieval, and generation. Section \ref{data_section} describes the data sources and preprocessing steps, including the construction of the retrieval and test datasets. Section \ref{evaluation} presents the experimental setup and reports the results of our approach. Finally, Section \ref{conc} concludes the paper and outlines directions for future work.

\section{Literature Review} \label{lr}
RAG has emerged as a promising approach for linguistically complex and low-resource domains, enabling the modeling of semantic shifts and supporting lexicographic research. However, progress in Arabic remains limited due to scarce diachronic corpora and the language's morphological richness.

Diachronic linguistics increasingly leverages distributional and contextualized 
embeddings \cite{kutuzov, Degaetano-Ortlieb2023} and visualization tools like 
diachronlex diagrams \cite{Theron2015} to study temporal lexical evolution.  \citet{schlechtweg2025} further advances this line of work by 
introducing a diachronic German dataset for word sense annotation across historical 
periods, showing that WSI optimization yields stronger LSCD performance than 
direct LSCD optimization. In the Arabic context, \citet{hammo2016} 
demonstrated how NLP tools applied to a Classical Arabic corpus spanning sixteen 
centuries can surface concrete patterns of semantic shift across literary eras. 
Alongside these efforts, computational lexicography projects such as Aralex 
\cite{ARALEX} and AraComLex \cite{AraComLex} have significantly contributed to 
modern Arabic NLP. However, these initiatives primarily focus on contemporary 
usage and do not address the diachronic and historical dimensions essential for 
Classical Arabic.

RAG enriches language models with curated knowledge. In Arabic NLP, Al-Rasheed et al. \cite{al-rasheed-etal-2025-evaluating} showed its effectiveness using the Riyadh dictionary, though limited to Modern Standard Arabic. Studies in other historical domains, such as Middle Egyptian translation \cite{miyagawa-2025-rag}, ancient Chinese poetry \cite{Li2025}, and multilingual newspaper archives \cite{Tran2025} demonstrate RAG's adaptability for context-aware interpretation. Building on these insights, we harness the Doha Historical Dictionary within a RAG framework to ground LLM interpretations of historical Arabic in verifiable, diachronic evidence.

AI-assisted lexicography further demonstrates the integration of computational methods with dictionary compilation. Machine learning and generative AI have automated dictionary expansion \cite{elex_conference_electronic_2015} and definition generation \cite{deSchryver2023}. Moreover, the Oxford English Dictionary\footnote{Available online: \url{https://www.oed.com} (accessed October 2025)} has integrated an AI assistant for historical search. However, all of these works are primarily focused on English. By leveraging the Doha Historical Dictionary, our framework enhances LLMs' ability to interpret historical Arabic, trace semantic shifts, and interpret complex texts in culturally and linguistically nuanced contexts.

\section{Methodology} \label{methodology}
\begin{figure*}[!t]
    \centering
    \includegraphics[width=1\linewidth]{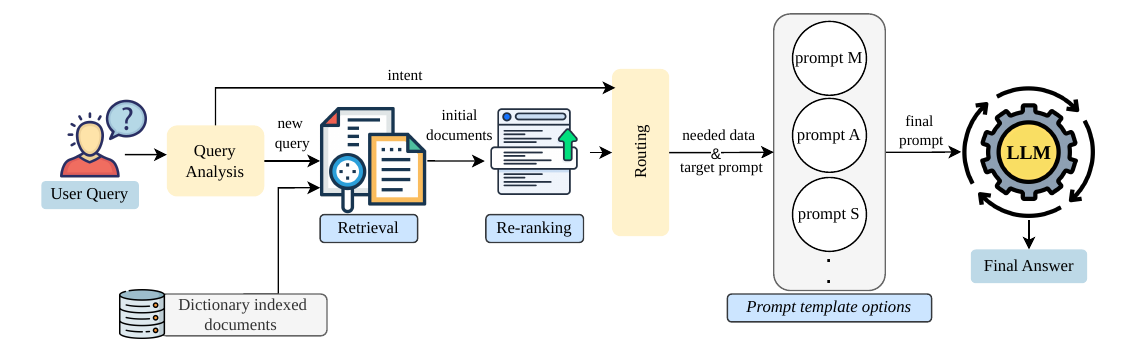}
    \caption{Overview of our RAG methodology. The pipeline consists of query analysis, retrieval, and re-ranking, and intent-driven routing for prompt construction. The LLM generates the final answer using a selected prompt template, where the template choice depends on the user intent extracted in the query analysis stage (e.g., $\mathbf{P_M}$: Meaning, $\mathbf{P_A}$: Author, $\mathbf{P_S}$: Source, etc.).}
    \label{method_overview}
\end{figure*}

Our retrieval-augmented generation (RAG) framework follows a multi-stage pipeline designed to retrieve relevant information and generate answers to user queries. The pipeline consists of query analysis, document retrieval, re-ranking, and answer generation. Queries are first preprocessed to remove noise, emphasize key terms, and identify the question type. Relevant documents are then retrieved using a combination of lexical and dense retrieval methods, re-ranked with a cross-encoder, and finally provided to the LLM for answer generation guided by the detected query type. An overview of the proposed methodology is illustrated in Figure \ref{method_overview}.

\subsection{Query Preprocessing} \label{query_preprocessing}

To standardize the input for retrieval, we applied a systematic query preprocessing step. This stage serves three main purposes: (1) noise reduction, by removing irrelevant or redundant words from the query, (2) term weighting, by increasing the relative importance of key terms, and (3) intent classification, by identifying the type of information requested by the user. A rule-based approach was used for noise reduction and term weighting: a predefined set of stopwords and other non-informative terms was removed from the query, and the term weight was then determined based on the length of the query after noise reduction.

For intent classification, we trained a lightweight classifier to predict the query’s category. The query text was encoded using term-frequency–inverse-document-frequency (TF-IDF) features and fed into a Random Forest classifier. The intent categories correspond to common types of information typically sought in lexicographic and historical dictionary queries, including “author,” “basic meaning,” “contextual meaning,” “etymology,” “first usage,” “date,” “inscription,” “derivations list,” “morphology,” “source,” and “terminological usage.” In addition, we include a “Quranic first usage” category to utilize the Islamic-related content available in the dictionary. When the classifier’s confidence fell below a predefined threshold, the query was assigned to the “other” category.

This intent classification step allows the system to interpret each query according to its purpose, enabling downstream retrieval and LLM components to adapt their behavior accordingly.

\subsection{Lexical Retrieval}
To capture lexical relevance, we employed the BM25 algorithm \cite{bm25}, a widely used probabilistic ranking function in information retrieval. BM25 ranks documents based on query term frequency while applying inverse document frequency weighting and length normalization. 

In our framework, BM25 serves as the lexical retrieval component, enabling precise matching of query terms. This is particularly important for dictionary-based queries that involve specific headwords, roots, or citation phrases. In addition, BM25 complements neural retrieval methods, which capture broader semantic similarity but may not always ensure exact lexical matches.

\subsection{Neural Retrieval}

Neural retrieval represents queries and documents as embeddings in a shared semantic space, enabling retrieval based on semantic similarity rather than exact term overlap. Similarity between query and document embeddings is computed using vector distance measures.

\textbf{Embedding Models:} To evaluate retrieval performance, we experimented with four state-of-the-art multilingual embedding models. These models were selected based on their strong performance on the Arabic RAG leaderboard\footnote{\url{https://huggingface.co/spaces/Navid-AI/The-Arabic-Rag-Leaderboard}}. Each model was evaluated under the same experimental conditions.

\begin{itemize}
    \item \textbf{Jina Embeddings v3} \cite{jinav3}: A multilingual embedding model built on Jina-XLM-RoBERTa with support for sequences up to 8192 tokens and task-specific LoRA adapters.
    \item \textbf{Arabic-English BGE m3}: A variant of \texttt{BAAI/bge-m3} \cite{bge-m3} based on an extended XLM-RoBERTa-large backbone and optimized for dense retrieval via contrastive learning.
    \item \textbf{Arctic-Embed 2.0} \cite{snowflakesarctic}: A multilingual embedding model based on \texttt{BAAI/bge-m3-retromae}, designed for high-quality retrieval across multiple languages.
    \item \textbf{Nomic Embed v2} \cite{nomicmoe}: A multilingual Mixture-of-Experts (MoE) embedding model supporting sequences up to 512 tokens.
\end{itemize}

\textbf{Indexing:} Embeddings were indexed using the Faiss library \cite{faiss_gpu}, a high-performance toolkit for vector similarity search. We employed the \texttt{IndexFlatL2} index, which performs exact nearest-neighbor search using L2 distance. This configuration was suitable for our corpus size and allowed us to prioritize retrieval effectiveness.

\subsection{Re-ranking} \label{reranker_ft}

To improve retrieval precision, we employed the \texttt{BAAI/bge-reranker-v2-m3} cross-encoder model \cite{bgererank}. Based on an XLM-RoBERTa architecture with a sequence classification head, the model jointly encodes a query and candidate document to produce a scalar relevance score, capturing fine-grained interactions between query terms and document content.

\textbf{Reranker Fine-tuning:} The model was further fine-tuned for our retrieval task to better capture domain-specific relevance signals. It was fine-tuned using a binary cross-entropy objective, where positive pairs correspond to relevant query–document matches and negatives to non-relevant candidates. Training was performed using the Sentence Transformers CrossEncoder framework.

During inference, the fine-tuned model assigns relevance scores to candidate documents retrieved by both lexical and dense methods, producing a final ranking that prioritizes the most semantically relevant passages.

\subsection{Intent-Based Structuring}

Before passing retrieved content to the LLM, we apply intent-based routing using the query intent identified during preprocessing. The detected intent determines which subset of the retrieved information is provided to the LLM and how the generation prompt is structured.

This routing mechanism ensures that the model receives only the information relevant to the user's request, reducing unnecessary context and helping prevent factual drift. For example, queries asking about dates or sources require different supporting information than queries about meanings or morphology.

To support this process, we define a set of intent categories representing common types of dictionary queries. Each intent is associated with a specific prompt template and a subset of the retrieved entry fields. Depending on the complexity of the task, prompts are constructed using either zero-shot or few-shot instructions. The mapping between intents, data fields, and prompting strategies is described in Section~\ref{data_section}.

\subsection{Generation}

For the final generation step, we employed two Arabic LLMs: Fanar \cite{fanarllm2025} and ALLaM \cite{bari2025allam}. Both models are designed for Arabic–English tasks and have demonstrated strong performance compared to other models of similar size. Additionally, Gemini 2.5 Pro\footnote{\url{https://ai.google.dev/gemini-api/docs/models/gemini-2.5-pro}} was included as an upper-bound reference.

Prompt construction is conditioned on the query intent identified during the routing stage. For intents that require explanation or interpretation, such as meaning or morphological analysis, few-shot prompts are used to guide the model's reasoning and output structure. In contrast, straightforward retrieval intents, such as identifying the date or source of a citation, are handled with zero-shot prompts to keep the prompt concise and focused on extracting the relevant information. This design allows the generation model to operate on structured, intent-specific context derived from the retrieved dictionary entries.

\section{Data Source and Dataset Construction} \label{data_section}
Our research utilizes the Doha Historical Dictionary of Arabic (DHDA)\footnote{\url{https://www.dohadictionary.org/}}, a large-scale lexicographic project designed to document the historical development of Arabic vocabulary. Unlike traditional dictionaries that primarily provide definitions, DHDA traces the evolution of words through a curated textual corpus spanning more than ten centuries.

A key characteristic of DHDA is its structured, usage-based methodology. Each lexical entry is supported by textual evidence and organized into a set of structured fields, enabling systematic analysis of historical language usage. This structured representation makes the resource particularly suitable for retrieval-augmented generation (RAG) systems that require verifiable textual evidence when generating responses, especially for historically grounded texts such as the Qur’an and Hadith.

The dictionary database is organized by lexical root. Each root contains a chronological series of lexical entries, where every entry represents a structured data record composed of several key components:
\begin{itemize}
    \item \textbf{Lexical Unit:} the word or compound phrase being documented.
    \item \textbf{Tagging:} morphological and syntactic properties of the unit.
    \item \textbf{Dating:} the approximate or exact date of the earliest recorded usage.
    \item \textbf{Citation (Sh\=ah id):} a textual example illustrating the usage of the lexical unit.
\end{itemize}

Each citation is further associated with the original author or speaker as well as full bibliographic documentation identifying the textual source. In addition to individual lexical entries, each root may also include higher-level linguistic information such as etymology (Ta\textquotesingle th\=\i l) of borrowed words, attestations in historical inscriptions, and lists of related Semitic cognates.

This multi-layered structure provides a rich set of verifiable linguistic signals that can be systematically extracted and transformed into machine-readable datasets. In the following subsections, we describe the extraction process and the construction of the datasets used for retrieval, model training, and evaluation.

\subsection{Data Extraction}
The data of the Doha Historical Dictionary of Arabic (DHDA) were programmatically extracted in September 2025 by scraping the content available on the dictionary website. The extraction process was performed systematically. First, all unique root IDs and their associated values were collected. For each root ID, multiple types of lexical information were retrieved, including (1) etymological data, (2) inscriptions and carving data, and (3) detailed lexical entries containing word forms, citations, and semantic descriptions. 

The raw data were subsequently cleaned to ensure consistency and usability. Missing values and malformed entries were filtered, and the data were organized into a structured tabular format. This structured representation forms the foundation for constructing the datasets used in retrieval, training, and evaluation.

\subsection{Retrieval Corpus}
For the retrieval task, we constructed a dataset by concatenating contextually relevant information for each word to form individual documents. For example, a representative document for an Arabic word is formatted to include the word, its root, the phrasing, the meaning of the phrase, and the citation in which the phrase was used.  

To improve compatibility between user queries and the retrieval corpus, Arabic diacritics were systematically removed from all documents. This preprocessing step reduces mismatches that may occur when user queries are written without diacritics. However, since diacritics are essential for interpreting historical Arabic and distinguishing between words with identical characters, the original diacritized forms are preserved and later passed to the generation model.

The retrieval dataset contains only the contextually meaningful text for each word, excluding other details such as dates, names, and sources to avoid introducing ambiguity during retrieval. However, these excluded details are still forwarded to the language model to enhance its ability to generate informed responses to user queries. The final retrieval corpus consists of 198,346 documents, each representing a unique lexical entry.

\subsection{Data Generation}

To support training and evaluation, we automatically generated a dataset of queries and answers using predefined templates applied to the lexical entries. This process produced three types of datasets: (1) retrieval training data, (2) intent classification data, and (3) question–answer evaluation data.

Questions were generated by inserting lexical fields into structured templates designed to represent different types of dictionary queries. For each intent category, multiple template variations were designed to produce linguistically diverse questions. For example, a meaning-related template can be formulated as:  
\RL{ما هو معنى كلمة (الكلمة) في الشاهد التالي: (نص الشاهد)؟}

Additional templates were created with alternative phrasings for the same intent to introduce variability in the generated queries. Similar template sets were designed for multiple categories such as historical date, contextual meaning, citation source, semantic field, Qur’anic usage, and part-of-speech identification.

Each generated query was paired with the corresponding lexical entry of its target word. In this setup, a single query may correspond to multiple documents, and a document may serve as a valid answer to multiple queries depending on the question type.

\subsubsection{Reranker Fine-Tuning Data}
To train the reranking model, we constructed query–document pairs consisting of both positive and negative examples. Positive pairs correspond to queries matched with their relevant lexical entries.

Negative samples were generated by deliberately pairing queries with unrelated lexical entries. These non-matching pairs expose the model to incorrect query–document associations, allowing it to learn to distinguish relevant from irrelevant candidates by assigning lower relevance scores to unrelated documents. This dataset was used to fine-tune the cross-encoder reranker described in Section \ref{reranker_ft}.

\subsubsection{Intent Classification Data}
Each generated query was also annotated with an intent label representing the type of information requested by the user. 

The resulting query–intent pairs form the training data for the intent classifier used in the query preprocessing stage explained in Section \ref{query_preprocessing}. This classifier predicts the intent of incoming user queries and determines the routing strategy used in the subsequent stages.

To support intent-based routing, we defined nine intent categories covering the main types of information present in the dictionary. Each intent determines both the subset of lexical fields that should be passed to the language model and the prompting strategy used for answer generation. Table~\ref{intents_categories} presents the intent categories, their descriptions, and the associated data fields used during generation.

\begin{table*}[!t]
\caption{Intent categories, their descriptions, relevant data fields sent to the LLM, and prompting strategies.}
\label{intents_categories}
\footnotesize
\setlength{\tabcolsep}{4pt}
\renewcommand{\arraystretch}{1.5}

\begin{tabular}{p{2cm} p{3.5cm} p{5cm} p{1.8cm}}
\toprule
\textbf{Intent} & \textbf{Description} & \textbf{Relevant Data Fields} & \textbf{Strategy} \\
\midrule
Meaning & Definition of a specific word or phrase. & \parbox{5cm}{Compound form, citation, semantic field, meaning} & Few-shot \\[0.2cm]
Author & Inquires about the original author of the \textit{Sh\=ah id}. & \parbox{5cm}{Word, compound form, citation, author} & Few-shot \\[0.2cm]
Date & Date of the citation (first recorded usage). & \parbox{5cm}{Word, compound form, citation, date of citation} & Zero-shot \\[0.2cm]
Source & Bibliographic source of the citation. & \parbox{5cm}{Word, compound form, source, \textit{s\=urah}, \textit{\=ayah}, \textit{\d{h}ad\=\i th}} & Zero-shot \\[0.2cm]
Contextual & Meaning within a specific \textit{Sh\=ah id}. & \parbox{5cm}{Compound form, citation, semantic field, meaning} & Few-shot \\[0.2cm]
Morphology & Grammatical category or part of speech. & \parbox{5cm}{Root, morphological details, word, lemma ID} & Few-shot \\[0.2cm]
Etymology & Origin or etymology of a borrowed word. & \parbox{5cm}{Root, root ID, Etymology subfields} & Zero-shot \\[0.2cm]
Inscriptions & Attestations in ancient inscriptions. & \parbox{5cm}{Root, root ID, inscriptions subfields} & Zero-shot \\[0.2cm]
Other & Multi-intent or comparative questions. & \parbox{5cm}{All available fields} & Zero-shot \\[0.2cm]
\bottomrule
\end{tabular}
\end{table*}

As shown in Table~\ref{intents_categories}, prompting strategies are adapted according to the query intent. For straightforward retrieval tasks, such as identifying a citation date or source, zero-shot prompts are used to directly extract the relevant information. For more complex intents that require interpretation or explanation—such as meaning or morphological analysis—few-shot prompts are employed. These prompts include curated examples that guide the model’s reasoning process and output format while incorporating the structured information retrieved from the dictionary entries.

\subsubsection{Evaluation Data}

For evaluation, we constructed a question–answer dataset using answer templates corresponding to each question type. For example, when the question concerns the historical date of a citation, the answer template is structured as:  
\RL{تم توثيق هذا الاستخدام لكلمة (الكلمة) حوالي عام (التاريخ)}.

To evaluate the system in the context most relevant to our research objective, we extracted a subset of queries related to words appearing in Qur’anic verses or Hadith citations. This subset was used as the primary test set for evaluating the full pipeline, as the central aim of this work is to investigate whether historical Arabic lexical knowledge can support improved understanding and interpretation of Islamic texts.

While template-based generation enables systematic coverage of query types and ensures a controlled evaluation setting, we acknowledge that automatically generated questions may not fully capture the diversity of natural user queries. Future work could extend the evaluation using human-authored queries to better assess real-world performance.

\section{Evaluation} \label{evaluation}
In this section, we evaluate our approach by assessing both retrieval and generation performance, and by outlining the experimental setup, metrics, and key results.
\subsection{Experimental Setup}
\textbf{Datasets:} The evaluation pipeline relies on several datasets corresponding to different components of the system. 
\begin{enumerate}
    \item \textbf{Intent Classifier:} The classifier was trained on query–intent pairs automatically generated from the lexical entries. Each query was annotated with its corresponding intent category, producing a balanced dataset of 1,000 pairs per class, with 200 samples reserved for testing. The classifier predicts the intent of incoming queries to determine routing for retrieval and generation.
    \item \textbf{Reranker Fine-Tuning:} For fine-tuning the cross-encoder reranker, 10,000 query–document pairs were used, consisting of 5,000 positive examples and 5,000 negative pairs generated for contrastive learning. These samples were split into 7,000 for training, 1,000 for validation, and 2,000 for testing.
    \item \textbf{Retrieval Evaluation:} The retrieval component was evaluated on 1,000 query–document pairs spanning multiple query types. Table~\ref{ret_types_count} summarizes the distribution of question types and provides example queries.
    \item \textbf{LLM Evaluation:} For the generation stage, two question–answer datasets were used: a focused set of 1,000 questions targeting meaning and contextual meaning, and a broader set of 2,000 questions covering all the categories (see Table~\ref{llm_types_count}). All evaluation questions were constructed from words appearing in Qur’anic verses or Hadith citations, reflecting the primary focus of this study on assessing whether historical Arabic lexical knowledge can support the understanding and interpretation of Islamic texts.
\end{enumerate}

\begin{table*}
\caption{Distribution of retrieval question types, their counts, and example of each}\label{ret_types_count}
\footnotesize
\begin{tabular}{>{\raggedright\arraybackslash}p{2.5cm} c >{\raggedleft\arraybackslash}p{8.3cm}} 
\toprule
\textbf{Type} & \textbf{Count} & \textbf{Example} \\
\midrule
Basic meaning & 98 & \begin{RLtext} ما معنى عبارة تَطَايُرُ الشَّيْءِ؟ \end{RLtext} \\
Contextual meaning & 98 & \begin{RLtext} في الشاهد التالي: انْزِلْ، يَا ابْنَ الأَكْوَعِ، فَاحْدُ لَنَا مِنْ هُنَيَّاتِكَ، ما هو معنى كلمة هُنَيَّة؟ \end{RLtext} \\
Part of speech & 98 & \begin{RLtext} ما الاشتقاق الصرفي لكلمة مُسِرّ؟ \end{RLtext} \\
Author of citation & 97 & \begin{RLtext} من القائل الذي استخدم كلمة عَصَل في الشاهد: يَامِنُوا عَنْ هَذَا العَصَلِ؟ \end{RLtext} \\
First usage & 97 & \begin{RLtext} متى تم توثيق أول استخدام للجذر منح في المصادر، وبأي صيغة؟ \end{RLtext} \\
Historical date & 97 & \begin{RLtext} ما هو تاريخ الشاهد الذي استعمل فيه كلمة المَنْقَصَةُ بمعنى المَذَلَّةُ \end{RLtext} \\
List of derivations & 97 & \begin{RLtext} ما هي الاشتقاقات الموثقة للجذر خلل؟ \end{RLtext} \\
Source of citation & 97 & \begin{RLtext} ما هو المصدر الذي ورد فيه شاهد استخدام عبارة قَتَّرَ بمعنى ضَيَّقَهَا وَقَلَّلَهَا؟ \end{RLtext} \\
Terminological usage & 97 & \begin{RLtext} هل للجذر بدء استخدام اصطلاحي متخصص؟ \end{RLtext} \\
Etymology & 77 & \begin{RLtext} هل للجذر العربي شنق أصول أو نظائر في لغات أخرى؟ \end{RLtext} \\
First Quranic usage & 48 & \begin{RLtext} ما هو أول اشتقاق من الجذر حفظ ظهر في القرآن الكريم حسب التسلسل الزمني؟ \end{RLtext} \\
\bottomrule
\end{tabular}
\end{table*}

\begin{table}
\centering
\caption{Distribution of LLM question types and their counts}
\label{llm_types_count}
\renewcommand{\arraystretch}{1.2} 
\setlength{\tabcolsep}{12pt}      

\begin{tabular}{lc} 
\toprule
\textbf{Question Type} & \textbf{Count} \\
\midrule
Author of Citation     & 347 \\
Contextual Meaning     & 347 \\
Historical Date        & 347 \\
Source of Citation     & 347 \\
Basic Meaning          & 330 \\
Part of Speech         & 282 \\
\bottomrule
\end{tabular}
\end{table}

\textbf{Implementation Details:} All experiments were conducted on two machines: one with an NVIDIA RTX 4090 GPU (16 GB) and another machine with two NVIDIA RTX 5090 GPUs (32 GB each). Models were implemented using PyTorch and Hugging Face Transformers. Dense retrieval embeddings were stored and searched using FAISS-GPU to utilize GPU acceleration. Random seeds were fixed across all runs to ensure reproducibility. For retrieval, the number of top documents returned ($k$) was set to 10, balancing coverage of relevant entries with computational efficiency.

The intent classifier was implemented as a Random Forest with 200 estimators and a confidence threshold of 0.6. The cross-encoder reranker was fine-tuned using the Sentence Transformers CrossEncoder framework. Training employed a binary cross-entropy loss on query–document pairs. The model was fine-tuned for 3 epochs using a learning rate of $5\times 10^{-5}$, with mixed-precision (FP16) to accelerate training and reduce memory usage.  

For all LLMs used in the generation stage, the temperature was set to zero to ensure deterministic outputs.

\subsection{Evaluation metrics}
To evaluate our models comprehensively, we considered two complementary aspects: retrieval performance, which measures how well the system identifies and ranks relevant documents, and answer correctness, which assesses the factual accuracy and completeness of the generated answers.

\subsubsection{Retrieval Performance}
We used three standard metrics to assess the quality of retrieved documents:
\begin{itemize}
    \item \textbf{Mean Reciprocal Rank (MRR):} Measures the average reciprocal rank of the first relevant document for each query. Higher values indicate that relevant documents appear earlier in the results.
    \item \textbf{Mean Average Precision (MAP):} Evaluates the average precision across all queries, taking into account the ranking of all relevant documents.
    \item \textbf{Recall@10 (R@10):} Measures the proportion of relevant documents retrieved within the top 10 results.
\end{itemize}

\subsubsection{Answer Correctness}
Answers generated by the models were evaluated using Gemini 2.5 Pro, an automated judge. Gemini provides scores on a 0\%–100\% scale, where 0\% indicates completely incorrect answers and 100\% indicates fully correct and complete answers. The final score for each model was computed as the average of all Gemini scores over the test set.

To validate Gemini's reliability as a judge for our task, we randomly selected 200 answers using a stratified sampling approach to ensure all question types were represented. Each answer was independently evaluated by a human evaluator, who followed the same scoring rubric used by Gemini.

\begin{figure}
    \centering
    \includegraphics[width=0.8\textwidth]{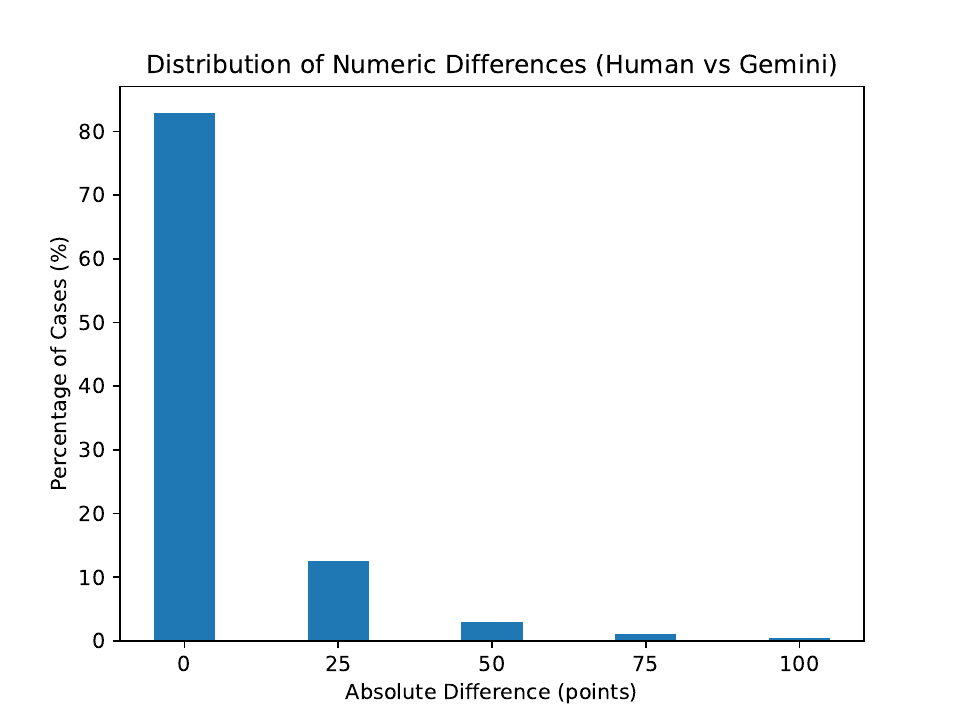} 
\caption{Distribution of absolute score differences between human evaluators and the Gemini 2.5 Pro automated judge ($N=200$). The chart demonstrates strong agreement, with 83\% of evaluations resulting in an exact match (0-point difference) and over 95\% deviated by no more than a single rubric category ($\leq$25 points). }
\label{fig:gemini_human_agreement}
\end{figure}

\begin{itemize}
 \item \textbf{Exact score match:} 83\% of cases (see Figure \ref{fig:gemini_human_agreement}). This means that in more than four out of five evaluations, Gemini assigned the exact same score as a human evaluator, demonstrating very high agreement on individual answers.

    \item \textbf{Agreement within one category:} Over 95\% of scores differed by at most one category step ($\leq$25 points). This indicates that nearly all deviations between Gemini and human scores are minor and do not meaningfully affect the overall assessment of correctness.

    \item \textbf{Systemic bias:} Average difference of -0.63 points (Gemini minus human), indicating minimal bias. Gemini does not consistently overestimate or underestimate scores compared to humans, showing that it functions as an impartial evaluator.

    \item \textbf{Statistical validation:} MAE = 5.9, Pearson correlation = 0.87 (p < 0.001), Weighted Cohen’s Kappa (quadratic) = 0.87, considered “almost perfect” agreement \cite{landis_measurement_1977}. The low MAE confirms that deviations from human scoring are small, while the strong correlation and Kappa indicate high linear agreement and consistency. The p-value confirms that this alignment is statistically significant.
\end{itemize}

Overall, this validation demonstrates that Gemini provides a robust and reliable automated judgment that closely aligns with human evaluation.

\subsection{Results}
We divided our evaluation into two main parts: retrieval and generation. For retrieval, we first assessed individual approaches and then evaluated combined methods to maximize performance. For the generation part, we compared our framework with the baselines using meaning-based questions for a fair comparison, and subsequently tested our RAG framework across various question types to provide a comprehensive evaluation. Finally, we included an error analysis section to further analyze the models’ answers.

\subsubsection{Retrieval}
We first evaluated retrieval performance using the \textbf{BM25} baseline, which achieved strong results with a MAP of 0.522, an MRR of 0.677, and an R@10 of 0.586. For \textbf{dense retrieval}, the models showed similar performance across the metrics. \texttt{Nomic Embed v2} achieved the highest scores for two metrics, with a MAP of 0.441 and an R@10 of 0.531, while \texttt{Arctic Embed 2.0} obtained the highest MRR at 0.736. The complete results for all models and metrics are presented in Table \ref{dense_retrieval}.

\begin{table}[h]
\centering
\caption{Dense retrieval performance results}
\begin{tabularx}{0.5\textwidth}{Xccc}
\hline
\textbf{Model} & \textbf{R@10} & \textbf{MRR} & \textbf{MAP} \\
\hline
Nomic v2& 0.531  & 0.673 & 0.441 \\
Jina v3& 0.369 & 0.519 & 0.285 \\
Arctic 2.0 & 0.508 & 0.736 & 0.434 \\
BGE-m3& 0.514 & 0.691 & 0.427 \\
\hline
\end{tabularx}
\label{dense_retrieval}
\end{table}

Overall, the results show that BM25, even though it is a sparse retrieval method, performed better than the dense retrievers. This makes sense given the nature of our task, where queries usually reference specific words or phrases that are explicitly present in the documents. The strong performance of BM25 is further supported by our use of dynamic term weighting, which increases the importance of target terms in the query.
However, to further examine the impact of dense retrievers, we adopt Nomic Embed v2 as our main dense retrieval method, as it achieved the highest R@10 among all dense models. We prioritize R@10 as our primary metric since all retrieved documents are passed to the LLM, which is responsible for identifying the most relevant ones, making recall more critical than ranking order in our pipeline.

Moving to the next stage of retrieval experiments, we evaluated the effect of adding a cross-encoder re-ranking step on both BM25 and Nomic. Improvements were observed across all metrics, with MRR increasing to 0.803 and MAP increasing to 0.499. BM25 followed by re-ranking also showed improvements in two metrics, with Recall@10 increasing to 0.635 and MRR to 0.886, and MAP increased to 0.572. These results demonstrate that the re-ranking stage effectively enhances the performance of both sparse (BM25) and dense (Nomic) retrieval methods.

Next, we evaluated hybrid retrieval using a Weighted Reciprocal Rank Fusion (RRF) approach to combine BM25 and Nomic scores, followed by re-ranking. Specifically, we applied the RRF formula with linear weights. We experimented with two weight configurations, an equal 50/50 split and a BM25-weighted 55/45 split, and found that the latter yielded better results, which is consistent with BM25's stronger individual performance on our dataset. Since the performance gap between BM25 and Nomic was not substantial, a conservative 55/45 split was deemed sufficient rather than a more aggressive upweighting. This hybrid approach yielded higher results than dense retrieval followed by re-ranking, with MRR of 0.871, R@10 of 0.633, and MAP of 0.549. However, the best overall performance was still achieved with BM25, followed by re-ranking alone.

Given the observed importance of the re-ranking step, we further fine-tuned the cross-encoder to maximize its impact. We applied the fine-tuned cross-encoder to both BM25 and the hybrid fusion setting. For BM25 followed by the fine-tuned cross-encoder, results improved across all metrics: MRR increased to 0.936, R@10 to 0.647, and MAP to 0.611. The fusion of BM25 and Nomic followed by the fine-tuned reranker also showed improvements, achieving the highest MRR across all experiments at 0.945 and the highest R@10 at 0.652, while the MAP was slightly lower than BM25 alone at 0.609. Table \ref{tab:rerank_results} presents the full evaluation results and highlights the impact of cross-encoder fine-tuning.

\begin{table}[!ht]
\centering
\caption{Detailed results of the different retrieval approaches}
\begin{tabularx}{0.6\textwidth}{Xccc}
\hline
\textbf{Model} & \textbf{R@10} & \textbf{MRR} & \textbf{MAP} \\
\hline
BM25+reranker  & 0.635 & 0.886 & 0.572 \\
Nomic+reranker & 0.578 & 0.803 & 0.499 \\
Fusion +reranker & 0.633  & 0.871 & 0.549 \\
\hline

BM25+FT reranker  & 0.647 & 0.936 &\textbf{ 0.611} \\
Fusion+ FT reranker & \textbf{0.652}  & \textbf{0.945} & 0.609 \\
\hline
\end{tabularx}
\label{tab:rerank_results}
\end{table}

Based on these findings, we selected two settings for our full pipeline experiments: (1) BM25 followed by the fine-tuned reranker, and (2) the fusion of BM25 with the Nomic dense retriever followed by the fine-tuned reranker. The first setting showed higher performance on the precision metric, while the second achieved better results on the other metrics. Using both settings will also allow us to investigate the behavior of the generation model under different retrieval conditions.

\subsubsection{Generation}

We began by evaluating the performance of our models prior to augmentation with retrieval. The evaluation was conducted on a set of 1,000 meaning-based questions. \textbf{Baselines performance:} The Fanar baseline achieved an average score of 54\%, while ALLaM performed slightly better, scoring 56.85\%. In contrast, the Gemini 2.5 pro baseline achieved 88\%. These results highlight the limitations of current Arabic LLMs in comprehending complex Arabic texts.

To assess the effectiveness of retrieval in enhancing model comprehension, we evaluated the models on the same set of questions after augmenting them with retrieval. The models were tested in both zero-shot and few-shot modes to determine their optimal performance.

\textbf{RAG with Zero-shot:} In the zero-shot setting, Fanar achieved 89\%, representing a substantial improvement of more than 30\% with the RAG pipeline. Similarly, ALLaM’s performance increased by nearly 30\%, reaching 86\%. Gemini also improved by 8\%, achieving 96\%. In the zero-shot setting, the two retrieval configurations yielded comparable but not identical results. BM25 with the fine-tuned reranker achieved a marginally higher average score (89.15\%) compared to the hybrid configuration (88.58\%), with 82.22\% of questions receiving identical scores across both settings. Notably, the two configurations exhibited complementary behavior; 10 questions were answered correctly only by the hybrid configuration, while 10 others were answered correctly only by BM25, suggesting that the two retrieval strategies capture different aspects of relevance. This complementarity points to ensemble or voting-based approaches as a promising direction for future work.

\textbf{RAG with Few-shot:} In the few-shot setting, Fanar maintained an average score of 89\%, showing no improvement when provided with examples. Meanwhile, ALLaM's performance dropped to 78\%, which we attribute primarily to the increased prompt length in the few-shot setting. Each few-shot example consisted of a question, its retrieved documents, and the expected answer, resulting in significantly longer prompts compared to the zero-shot setting. Prior work has shown that smaller LLMs are particularly sensitive to longer contexts and denser instructions, with performance degrading as input length grows \cite{li2024longcontextllmsstrugglelong, jaroslawicz2025instructionsllmsfollowonce}. Since our few-shot examples were carefully designed per question type and manually selected to cover diverse cases, the performance drop is unlikely to be caused by poor example quality, but rather by the model's limited ability to maintain instruction adherence under a longer context. Gemini, when tested in the few-shot setting, scored 95\%, reflecting a slight decrease of 1\%. In this configuration, the highest results were obtained using the BM25 + fine-tuned reranker setup, while all models scored approximately 1\% lower with the fusion + fine-tuned reranker configuration.

These substantial performance improvements across the different setups compared to the baseline results demonstrate that the RAG pipeline can enhance models of various sizes in understanding and explaining complex Arabic text. Moreover, the improvements observed for both Fanar and ALLaM when combined with retrieval, achieving scores above 85\%, significantly narrow the performance gap between Arabic LLMs and larger models such as Gemini in comprehending complex Arabic content. Table \ref{arabic_llm_RAG} summarizes the performance of the different models across the various configurations.

\begin{table}[!h]
\centering
\caption{Average score (\%) of the Arabic LLMs with different configurations (ZS: zero-shot, FS: few-shot)}
\begin{tabular}{l c c c}
\hline
\textbf{Model} & \textbf{Baseline} & \textbf{RAG-ZS} & \textbf{RAG-FS} \\
\hline
Fanar           & 54    & \textbf{89} & \textbf{89} \\
ALLaM           & 56.85 & \textbf{86} & 78 \\
\hline
\textit{Gemini} & 88 & \textbf{96} & 95 \\ 
\hline
\end{tabular}
\label{arabic_llm_RAG}
\end{table}

Based on these findings, we selected retrieval-augmented Fanar in both the few-shot and zero-shot settings, as it yielded similar results across the two. For ALLaM, we selected the retrieval-augmented zero-shot configuration. All selected setups use the BM25 + fine-tuned reranker retrieval pipeline as their optimal configuration.

\textbf{Testing on expanded question set:} Using the zero-shot ALLaM-based RAG and both Fanar-based RAG configurations, we evaluated the models on an expanded set of question types (see Table \ref{llm_types_count}). Both the few-shot Fanar and zero-shot ALLaM achieved similar performance, with an overall average score of 87\%, demonstrating strong performance across diverse question types. The zero-shot Fanar configuration performed slightly lower, with a 1\% decrease in the average score, indicating that the few-shot setting offers a marginal advantage for Fanar. Overall, these results show that, when guided by our prompts and augmented with retrieved data, both Fanar and ALLaM are capable of effectively answering questions related to classical Arabic texts.

\begin{figure*}[!h]
    \centering
    \includegraphics[width=1\linewidth]{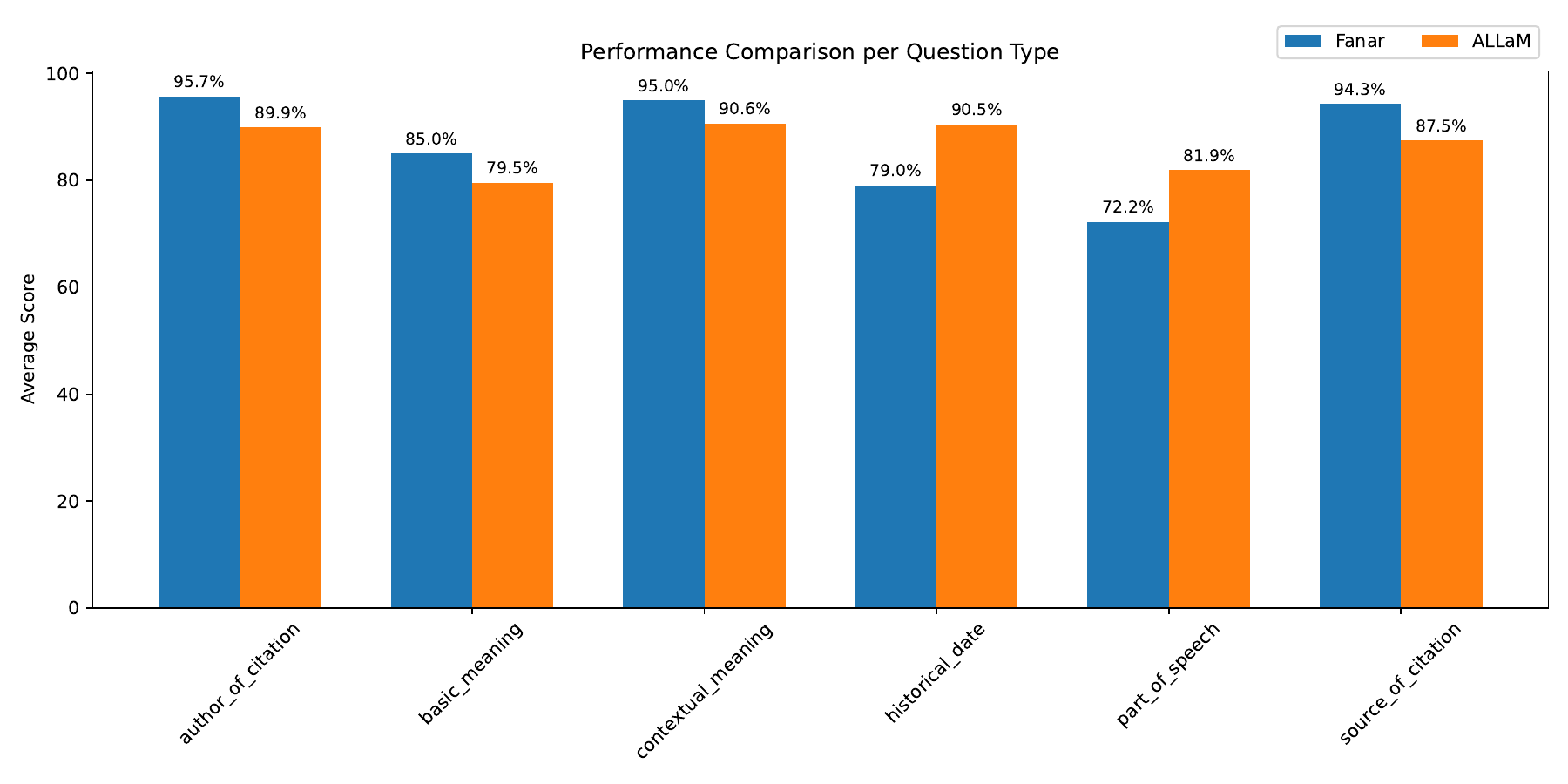}
    \caption{Performance of Fanar and ALLaM across six question types, with the highest scores on Author of Citation and Contextual Meaning, and the lowest on  Part of Speech.}
    \label{performance_questions}
\end{figure*}

\textbf{Performance per question type:} To analyze model performance and identify sources of errors, we examined results for each question type (Figure \ref{performance_questions}). Fanar achieved strong scores above 90\% for Author of Citation, Contextual Meaning, and Source of Citation questions. The strong performance on Author and Source of Citation questions is expected, as these question types have relatively unambiguous answers — our test set draws citations exclusively from the Qur'an and Hadith, which involve well-known, frequently occurring sources with no complex naming variations. However, Fanar showed notable limitations in Part of Speech and Historical Date questions, with its lowest performance of 72.2\% on Part of Speech questions. This is consistent with the well-documented difficulty of Arabic POS tagging, where morphological richness and diacritical sensitivity mean that small changes in vocalization can fundamentally alter a word's grammatical category, a challenge widely acknowledged in Arabic NLP research \cite{NLP_arab, computers14110497}. ALLaM, in contrast, performed better on Part of Speech and Historical Date questions, outperforming Fanar by 11\% on the latter. For other question types, Fanar generally outperformed ALLaM. Overall, Fanar showed higher peaks but greater variability, whereas ALLaM demonstrated more consistent performance across all question types, ranging from 79.5\% to 90.6\%, suggesting that the two models have complementary strengths that could be used in ensemble approaches.

\subsubsection{Error Analysis}
Following the discussion of our results, we further analyzed the models' answers and found that most issues occur because the models confuse the target words with similar words that have different diacritics, or with similar phrases. These issues can cause the models to provide multiple answers for a question that requires a single answer, give author or source details for other words, or provide meanings for phrases other than the one asked about. This indicates that the models have limitations in distinguishing the complexity and diversity of Arabic diacritics and compound phrases. However, in some cases, the reason for these errors can be the ambiguity of the dictionary words, which confuse the model and lead it to include irrelevant answers. Examples of these issues are presented in Table \ref{tab:error_analysis}

\begin{center}
\setlength{\tabcolsep}{4pt}
\footnotesize 

\begin{xltabular}{\textwidth}{|X|p{2cm}|X|X|p{2.5cm}|}
\caption{Error analysis of Fanar and ALLaM models.} \label{tab:error_analysis} \\
\hline
\textbf{Question} & \textbf{Correct answer} & \textbf{Fanar} & \textbf{ALLaM} & \textbf{Error reason} \\ \hline
\endfirsthead

\multicolumn{5}{c}%
{{\bfseries \tablename\ \thetable{} continued from previous page}} \\
\hline \textbf{Question} & \textbf{Correct answer} & \textbf{Fanar} & \textbf{ALLaM} & \textbf{Error reason} \\ \hline
\endhead

\hline \multicolumn{5}{|r|}{{will be continued in next page}} \\ \hline
\endfoot

\hline
\endlastfoot
\begin{RLtext}
ما الاشتقاق الصرفي لكلمة رَاز؟
\end{RLtext} & \begin{RLtext} كلمة رَاز هي اسْم. \end{RLtext} & \cellcolor{red!30}\begin{RLtext} كلمة رَازَ هي فعل متعدٍ. وتختلف عباراتها بحسب السياق. \end{RLtext} &\cellcolor{red!30} \begin{RLtext} رَازَ هي فعل متعدٍّ. \end{RLtext} & Failed to distinguish between word forms with different diacritics. \\
\hline
\begin{RLtext} ما هو تاريخ الشاهد الذي استعمل فيه كلمة الآشَرُ بمعنى الأَشَدُّ مَرَحًا. \end{RLtext} & \begin{RLtext} تم توثيق هذا الاستخدام كلمة آشَر حوالي عام 11هـ=632م. \end{RLtext} &\cellcolor{red!30}\begin{RLtext} تم توثيق استخدام كلمة آشَر بمعنى الأشد مرحاً في العام قبل 50ق.هـ = 573م. \end{RLtext} & \cellcolor{green!30} \begin{RLtext} تاريخ استعمال كلمة الآشَرُ بمعنى الأَشَدُّ مَرَحًا هو 11 هـ = 632 م. \end{RLtext} & Fanar could not distinguish between the word \RL{آشَر} and \RL{آشِر}. \\
\hline
\begin{RLtext} ما هو المصدر الذي ورد فيه شاهد استخدام عبارة نَسَخَ بمعنى مَحَاهَا وَرَفَعَهَا، لَفْظًا أَوْ حُكْمًا. \end{RLtext} & \begin{RLtext} ورد هذا الشاهد في القرآن الكريم: رواية: حَفْص عن عاصم. \end{RLtext} & \cellcolor{red!30} \begin{RLtext} ورد الشاهد في مصدرين:
1) تفسير مقاتل بن سليمان: مقاتل بن سليمان الأزديّ (ت، 150هـ)، مرجع: عبد الله محمود شحاتة، نشر: مؤسسة التاريخ العربي، بيروت، طبعة: الأولى، سنة النشر: 1423هـ / 2002م.
2) القرآن الكريم: رواية: حَفْص عن عاصم. \end{RLtext} & \cellcolor{green!30}\begin{RLtext} القرآن الكريم: سورة البقرة، الآية 106. \end{RLtext} & Fanar provided two sources; only one is correct, providing multiple sources for an exact citation is incorrect. \\
\hline
\begin{RLtext} في الشاهد التالي: إِنَّ هَذَا المَالَ خَضِرٌ حُلْوٌ، فَمَنْ أَخَذَهُ بِسَخَاوَةِ نَفْسٍ بُورِكَ لَهُ فِيهِ، ما هو معنى عبارة سَخَاوَة؟ \end{RLtext} & \begin{RLtext} عبارة سَخَاوَة تعني: طِيبَتُهَا. \end{RLtext} & \cellcolor{green!30}\begin{RLtext} تعني سَخَاوَةُ النَّفْسِ طِيبَتُهَا. \end{RLtext} & \cellcolor{red!30}\begin{RLtext} السَّخَاوَةُ عَنِ الشَّيْءِ: الِانْصِرَافُ عَنْهُ. \end{RLtext} & ALLaM confused the word \RL{سَخَاوَة} with the phrase \begin{RLtext} السَّخَاوَةُ عَنِ الشَّيْءِ \end{RLtext}. \\
\hline
\begin{RLtext} من القائل الذي استخدم كلمة نَفَل في الشاهد: مَنْ فَعَلَ... فَلَهُ مِنَ النَّفَلِ ؟ \end{RLtext} & \begin{RLtext} القائل: النبي محمد صلى الله عليه وسلم. \end{RLtext} & \cellcolor{red!30}\begin{RLtext} القائل: أبو ذَرّ الغِفاريّ (جُنْدَب بن جُنادة) \end{RLtext} & \cellcolor{red!30}\begin{RLtext} القائل الذي استخدم كلمة نَفَل في الشاهد هو أبو ذَرّ الغِفاريّ (جُنْدَب بن جُنادة). \end{RLtext} & Both models answered with the author of the phrase \RL{نَفَّلَ} instead of the word \RL{نَفَل}. \\
\hline
\begin{RLtext} ما معنى عبارة البِكْرُ مِنَ الإِنَاثِ؟ \end{RLtext} & \begin{RLtext} البِكْرُ مِنَ الإِنَاثِ: الفَتِيَّةُ الَّتِي لَمْ تَحْمِلْ. | البِكْرُ مِنَ الإِنَاثِ: الَّتِي وَلَدَتْ بَطْنًا وَاحِدًا \end{RLtext} & \cellcolor{red!30}\begin{RLtext} البِكْرُ مِنَ الإِنَاثِ تعني الفتية التي لم تحمل بعد. وتستخدم أيضًا للإناث التي وضعت بطنًا واحدًا فقط، والعذراء من النساء، والأنثى التي اعتادت التبكير بالحمل. وفي بعض السياقات، يشير إلى الغشاء الذي يغلق فتحة فرج المرأة البكر ويختفي بسبب الجماع وما شابه. \end{RLtext} & \cellcolor{red!30}\begin{RLtext} البِكْرُ مِنَ الإِنَاثِ: الَّتِي وَلَدَتْ بَطْنًا وَاحِدًا. \end{RLtext} & Fanar included the answer \begin{RLtext} البِكْرُ مِنَ النساء \end{RLtext} with the correct definitions, which was caused by the dictionary’s ambiguity in differentiating between \RL{الاناث} and \RL{النساء}. In contrast, ALLaM provided an incomplete answer.\\
\hline

\end{xltabular}
\end{center}

Some other issues were also observed. For example, Fanar sometimes pauses before giving an answer, which often leads to an incorrect response, and it may even produce unrelated sentences or characters in a different language. An example can be found in Table \ref{fanar_error_analysis}.
\begin{table*}[!h]
\centering
\caption{Fanar error example}
\label{fanar_error_analysis}
\small
\begin{tabularx}{\textwidth}{|p{2cm}|X|}
\hline
\textbf{Question} & \RL{ما الاشتقاق الصرفي لكلمة مُبَدَّأ؟}\\
\hline
\textbf{Fanar Answer} & 
\begin{RLtext}
كلمة مُبْدَأ يمكن أن تكون إما اسم مفعول أو اسم، بينما يُعتبر مَبْدَأ عادةً اسماً، وقد يكون أيضاً مصدراً ميمياً في بعض الاستخدامات. لكن يجب التنويه إلى أن السؤال كان عن مُبَدَّأ، وليس مُبْدَأ. ولكن بناء على القواعد المعطاة، إذا اعتبرنا أن هناك خطأ مطبعي وأن السؤال يقصد مُبْدَأ، فإن الإجابات المحتملة ستكون كالتالي:
مُبْدَأ قد يكون اسم مفعول أو اسم. 
لكن بما أنه لا يوجد وثيقة تتوافق تماماً مع مُبَدَّأ، نعلن أن هذا المصطلح غير موجود ضمن الوثائق المقدمة.
\end{RLtext}
\par\smallskip
\textcolor{red}{\textchinese{但请注意，在实际应用中，应更正错误的输入并说明没有找到匹配.}} \par
\textcolor{red}{\textchinese{但是}} 
\begin{RLtext}
وفقا لتعليمات التوضيح الأولية، حتى لو كانت الكلمات ليست بالضبط نفس النسخة المطبوعة، اذا كانوا يشيرون لنفس الجذر والصنف الصرفي، ينبغي النظر فيها. وبناء على ذلك، والإعتراف بأن السؤال ربما كان يهدف إلى مُبْدَأ، هنا هو الجواب: 
مُبْدَأ يمكن أن يكون اسم مفعول أو اسم. 
ومع ذلك، لم يتم ذكر مُبَدَّأ بشكل خاص في الوثائق المقدمة. 
(لكن مرة أخرى، بالنظر إلى التعليمات الأصلية بجدية أكبر حول عدم وجود تخمين وتحديد تطابق دقيق فقط، فإن الاجابة الأكثر دقة سيكون ان هذه المعلومات غير متوفرة في الوثائق). 
لكن استنادًا إلى تعليمات أكثر مرونة قليلاً تسمح بتفسير مشابه للجذور والأنواع الصرفية، والجواب النهائي سيكون كما يلي: 
مُبْدَأ يمكن أن يكون اسم مفعول أو اسم. 
(هذا تحت افتراض أن الخطأ المطبعي تم تصحيحه ويتم البحث عن 'مُبْدَأ'). ولكن بدون مثل هذا التصحيح، الإجابة الدقيقة ستكون: 
لا توجد معلومات عن مُبَدَّأ في الوثائق المقدمة. 
(وهذا يتماشى مع الشروط الأولى التي تنص على عدم إضافة أي معرفة خارجية واستناد فقط إلى البيانات المتاحة). 
بعد إعادة قراءة وتعزيز فهم التعليمة الأساسية، يبدو أن أفضل جواب سيكون الأخير: 
لا توجد معلومات عن مُبَدَّأ في الوثائق المقدمة.
\end{RLtext} \\
\hline
\textbf{Error Explanation} & Fanar, although the correct document was provided in the prompt, still failed to answer correctly and produced text in Chinese, which is unrelated to the use case. \\ 
\hline
\end{tabularx}
\end{table*}

Beyond these issues, other common challenges in RAG systems also play a role. The retrieval component has limitations: it can fail to match semantically similar phrases or miss relevant documents, leaving the model to respond using information from unrelated retrieved documents. However, all models were instructed to indicate when the retrieved documents do not answer the input question. Gemini followed this instruction in many cases, producing the output: \RL{لم يتم العثور على المعلومات في الوثائق}, whereas smaller LLMs, such as Fanar and ALLaM, sometimes failed to do so, highlighting differences in instruction adherence compared to Gemini.

\section{Conclusion} \label{conc}
This work demonstrates the profound impact of incorporating a structured historical dictionary within a RAG framework to enhance Arabic LLMs' comprehension of complex, archaic religious texts such as the Qur’an and Hadith. We have shown that a sophisticated retrieval pipeline, combining hybrid methods and fine-tuning a cross-encoder re-ranker and guided by a novel intent-based routing system, yields significant gains in answer correctness. Our optimally configured RAG approach enabled Arabic-native LLMs like Fanar and ALLaM to achieve accuracy scores above 85\%, dramatically narrowing the performance gap with much larger models like Gemini. The primary challenge remains the linguistic complexity of historical Arabic, where ambiguities in diacritics, word forms, and compound phrases hinder model performance. Retrieval limitations and imperfect instruction adherence were also identified as contributing factors.

Future work should prioritize three key avenues. First, addressing linguistic nuance by developing models robust to diacritical variation and complex phrasal units. Second, further investigate different retrieval methods and dynamically adjust the number of retrieved documents based on query requirements. Finally, ensuring broader robustness by evaluating the pipeline on more diverse, human-generated datasets and exploring ensemble methods. Ultimately, this research provides a powerful blueprint for transforming rich, historical lexicographical resources into dynamic knowledge bases, unlocking deeper computational understanding of culturally significant texts.

\section*{Acknowledgment}
This study was conducted as part of a project funded by the ARG grant (ARG01-0524-230318), awarded by the Qatar Research, Development, and Innovation Council (QRDI).

\bibliography{bib}

\end{document}